\PassOptionsToPackage{unicode}{hyperref}
\PassOptionsToPackage{hyphens}{url}
\PassOptionsToPackage{dvipsnames,svgnames,x11names}{xcolor}
\documentclass[
  a4paper,
  12pt]{article}
\usepackage{xcolor}
\usepackage[top=15mm,bottom=17mm,left=16mm,right=16mm]{geometry}
\usepackage{amsmath,amssymb}
\usepackage{iftex}
\ifPDFTeX
  \usepackage[T1]{fontenc}
  \usepackage[utf8]{inputenc}
  \usepackage{textcomp} 
\else 
  \usepackage{unicode-math} 
  \defaultfontfeatures{Scale=MatchLowercase}
  \defaultfontfeatures[\rmfamily]{Ligatures=TeX,Scale=1}
\fi
\usepackage{lmodern}
\ifPDFTeX\else
\fi
\IfFileExists{upquote.sty}{\usepackage{upquote}}{}
\IfFileExists{microtype.sty}{
  \usepackage[]{microtype}
  \UseMicrotypeSet[protrusion]{basicmath} 
}{}
\makeatletter
\@ifundefined{KOMAClassName}{
  \IfFileExists{parskip.sty}{%
    \usepackage{parskip}
  }{
    \setlength{\parindent}{0pt}
    \setlength{\parskip}{6pt plus 2pt minus 1pt}}
}{
  \KOMAoptions{parskip=half}}
\makeatother
\usepackage{color}
\usepackage{fancyvrb}

\DefineVerbatimEnvironment{Highlighting}{Verbatim}{commandchars=\\\{\}}
\newenvironment{Shaded}{}{}

\newcommand{\ExtensionTok}[1]{#1}

\newcommand{\NormalTok}[1]{#1}

\usepackage{longtable,booktabs,array}
\usepackage{caption}
\usepackage{calc} 
\usepackage{etoolbox}
\makeatletter
\patchcmd\longtable{\par}{\if@noskipsec\mbox{}\fi\par}{}{}
\makeatother
\IfFileExists{footnotehyper.sty}{\usepackage{footnotehyper}}{\usepackage{footnote}}
\makesavenoteenv{longtable}
\providecommand{\tightlist}{%
  \setlength{\itemsep}{0pt}\setlength{\parskip}{0pt}}
\usepackage{microtype}
\usepackage[scale=0.88]{sourcecodepro}
\usepackage{longtable}
\usepackage{booktabs}
\usepackage{array}
\usepackage{colortbl}
\usepackage{titlesec}
\usepackage{titling}
\usepackage[most]{tcolorbox}
\usepackage{setspace}
\usepackage{enumitem}
\usepackage{ragged2e}
\usepackage{graphicx}
\usepackage{multicol}
\usepackage{needspace}
\usepackage{hyperref}

\definecolor{MathdownPaper}{HTML}{FFFFFF}
\definecolor{MathdownInk}{HTML}{24231F}
\definecolor{MathdownBody}{HTML}{34332E}
\definecolor{MathdownMuted}{HTML}{78756D}
\definecolor{MathdownLine}{HTML}{D8D3C8}
\definecolor{MathdownAccent}{HTML}{B84F32}
\definecolor{MathdownAccentSoft}{HTML}{F1DFD8}
\definecolor{MathdownEditor}{HTML}{F3F0E8}
\pagecolor{MathdownPaper}
\color{MathdownBody}

\setlist{topsep=0.55em,itemsep=0.34em,parsep=0pt,leftmargin=1.75em}
\setlist[enumerate]{label=\textcolor{MathdownAccent}{\arabic*.}}
\setlist[itemize]{label=\textcolor{MathdownAccent}{\textbullet}}
\renewcommand{\arraystretch}{1.22}
\arrayrulecolor{MathdownLine}
\AtBeginEnvironment{longtable}{\arrayrulecolor{MathdownLine}}

\titleformat{\section}
  {\sffamily\bfseries\color{MathdownInk}\fontsize{18}{22}\selectfont%
   \hyphenpenalty=10000\exhyphenpenalty=10000\relax}
  {}{0pt}{}
  [\vspace{0.18em}{\color{MathdownAccent}\titlerule[1.25pt]}]
\titleformat{\subsection}
  {\sffamily\bfseries\color{MathdownInk}\fontsize{14}{17}\selectfont}
  {}{0pt}{}
\titleformat{\subsubsection}
  {\sffamily\bfseries\color{MathdownInk}\fontsize{11.5}{14}\selectfont}
  {}{0pt}{}
\titlespacing*{\section}{0pt}{2.25em}{0.9em}
\titlespacing*{\subsection}{0pt}{1.7em}{0.55em}
\titlespacing*{\subsubsection}{0pt}{1.25em}{0.4em}

\newtcolorbox{paperabstract}{
  enhanced,
  breakable,
  colback=MathdownAccentSoft,
  colframe=MathdownAccentSoft,
  boxrule=0pt,
  borderline west={2.4pt}{0pt}{MathdownAccent},
  arc=2.5mm,
  left=5mm,
  right=5mm,
  top=24mm,
  bottom=23mm,
  extras first={bottom=6mm},
  extras middle={top=6mm,bottom=6mm},
  extras last={top=6mm},
  before skip=5mm,
  after skip=3.6em
}
\newcommand{\paperabstracttitle}{%
  {\sffamily\bfseries\color{MathdownAccent}\fontsize{17}{20}\selectfont Abstract}%
  \par\vspace{0.5em}}

\newtcolorbox{articletoc}{
  enhanced,
  colback=MathdownEditor,
  colframe=MathdownLine,
  boxrule=0.7pt,
  arc=2.5mm,
  left=5mm,
  right=5mm,
  top=4mm,
  bottom=4mm,
  before skip=5mm,
  after skip=0pt
}
\renewenvironment{Shaded}
  {\begin{tcolorbox}[
    enhanced,
    colback=MathdownEditor,
    colframe=MathdownLine,
    boxrule=0.7pt,
    arc=2.5mm,
    left=4mm,
    right=4mm,
    top=3mm,
    bottom=3mm,
    before skip=0.8em,
    after skip=1em]}
  {\end{tcolorbox}}
\newcommand{\toclabel}{%
  {\sffamily\bfseries\color{MathdownAccent}\fontsize{9}{11}\selectfont%
  \MakeUppercase{In this article}}\par\vspace{0.65em}}
\newcommand{\tocentry}[2]{%
  \par\noindent\hyperref[#1]{\color{MathdownInk}#2}\par\vspace{0.16em}}

\makeatletter
\renewcommand{\maketitle}{%
  \thispagestyle{empty}
  {\sffamily\bfseries\color{MathdownAccent}\fontsize{9}{11}\selectfont%
    \MakeUppercase{AI research}\par}
  \vspace{-1.55em}
  {\sffamily\bfseries\color{MathdownInk}\fontsize{27}{30}\selectfont%
    \raggedright\@title\par}
  \vspace{1.4em}
  {\color{MathdownLine}\hrule height 0.7pt}
  \vspace{1em}
  {\sffamily\fontsize{9.5}{12}\selectfont\color{MathdownMuted}%
    \textbf{\color{MathdownInk}Anand Murugan}%
    \hspace{0.9em}\textcolor{MathdownLine}{\textbullet}\hspace{0.9em}%
    August 17, 2026%
    \hspace{0.9em}\textcolor{MathdownLine}{\textbullet}\hspace{0.9em}%
    Research article%
    \hspace{0.9em}\textcolor{MathdownLine}{\textbullet}\hspace{0.9em}%
    A causal test of the LM-head bottleneck claim\par}
  \vspace{1em}
  {\color{MathdownLine}\hrule height 0.7pt}
  \vspace{7mm}
  \begin{articletoc}
    \toclabel
    {\sffamily\fontsize{10.5}{13}\selectfont
    \begin{minipage}[t]{0.48\linewidth}
      \tocentry{main-findings}{Main Findings}
      \tocentry{what-lost-means-here}{What ``Lost'' Means Here}
      \tocentry{a-clean-test-of-backward-rank}{A Clean Test of Backward Rank}
      \tocentry{compact-byte-level-language-model}{Compact Byte-Level Language Model}
      \tocentry{larger-subword-language-model}{Larger Subword Language Model}
      \tocentry{is-the-remaining-logit-error-already-used}{Is the Remaining Logit Error Already Used?}
      \tocentry{spamlang-repeated-tokens-are-not-independent-mapping-examples}{SpamLang: Repeated Tokens Are Not Independent Mapping Examples}
      \tocentry{increasing-the-output-vocabulary-without-changing-the-text}{Increasing the Output Vocabulary Without Changing the Text}
    \end{minipage}\hfill
    \begin{minipage}[t]{0.48\linewidth}
      \tocentry{do-projection-measurements-predict-learning}{Do Projection Measurements Predict Learning?}
      \tocentry{secondary-evidence-tested-auxiliary-feedback-routes}{Secondary Evidence: Tested Auxiliary Feedback Routes}
      \tocentry{what-the-experiments-establish}{What the Experiments Establish}
      \tocentry{relation-to-earlier-work}{Relation to Earlier Work}
      \tocentry{limits-of-this-study}{Limits of This Study}
      \tocentry{reproducing-the-results}{Reproducing the Results}
      \tocentry{conclusion}{Conclusion}
    \end{minipage}}
  \end{articletoc}
}
\makeatother

\AtBeginDocument{%
  \RaggedRight
}
\usepackage{bookmark}
\IfFileExists{xurl.sty}{\usepackage{xurl}}{} 
\makeatletter
\@ifundefined{xmpquote}{}{}
\makeatother
\hypersetup{
  pdftitle={Does the LM Head Create a Harmful Gradient Bottleneck? A Causal Test},
  pdfauthor={Anand Murugan anand.murugan@gmail.com},
  colorlinks=true,
  linkcolor={MathdownAccent},
  filecolor={Maroon},
  citecolor={MathdownAccent},
  urlcolor={MathdownAccent},
  pdfcreator={LaTeX via pandoc}}

\title{Does the LM Head Create a Harmful Gradient Bottleneck? A Causal
Test}
\author{Anand Murugan\\
\href{mailto:anand.murugan@gmail.com}{\nolinkurl{anand.murugan@gmail.com}}}
\date{August 2026}

\begin{document}
\maketitle

\begin{paperabstract}\paperabstracttitle

The language-model head maps a hidden state of width \(D\) to a
vocabulary of size \(V\), so its transpose can return at most \(D\)
independent directions to the Transformer. Godey and Artzi argue that
this severe projection is a harmful optimization bottleneck. We separate
the geometry from the causal claim. Our backward-only intervention keeps
the ordinary logits and the exact LM-head parameter update while
reducing only the rank of the gradient sent into the Transformer. Across
five paired seeds on byte-level and BPE-8192 WikiText-2 models, reducing
backward rank increases validation loss. An equally ranked factorized
forward head, however, increases loss substantially more. At half rank
in the larger model, the backward-only loss increase is \(0.0586\) (95\%
CI \([0.0167, 0.1005]\)), while the factorized forward head increases
loss by \(0.1795\) (\([0.1547, 0.2042]\)). The vocabulary-space residual
also contributes to the ordinary LM-head update, and removing that
contribution is harmful. Additional controls show that repeated-token
failures are confounded by the number of independently sampled symbols,
that adding never-target output classes does not impair learning, and
that projection diagnostics do not reliably predict progress in our
runs. Tested auxiliary feedback routes do not beat tuned
backpropagation. These results confirm strong geometric compression but
do not establish that it is a harmful optimization bottleneck.

\end{paperabstract}

The language-model head, or \textbf{LM head}, maps the
Transformer\textquotesingle s final hidden state to one logit for every
token in the vocabulary. For hidden width \(D\) and vocabulary size
\(V\), the standard head is a trainable matrix
\(W \in \mathbb{R}^{V \times D}\):

\[
z = Wh,
\]

where \(h \in \mathbb{R}^{D}\) and \(z \in \mathbb{R}^{V}\).
Backpropagation sends the logit gradient back to the hidden state
through:

\[
g_h = W^T g_z.
\]

The hidden-state gradient therefore depends only on the component of
\(g_z\) in the column space of \(W\), a subspace of dimension at most
\(D\). When \(V \gg D\), most of the Euclidean norm of \(g_z\) can lie
outside that subspace.

\href{https://arxiv.org/abs/2603.10145}{Godey and Artzi} call this a
gradient bottleneck. Their paper, \emph{Lost in Backpropagation: The LM
Head is a Gradient Bottleneck}, reports that the return path often
retains only 1--5\% of the logit-gradient norm. In other words, 95--99\%
lies outside the directions that can immediately reach the final hidden
state.

That geometric fact is real. The open question is what it means.

A large projection residual does not by itself show that useful learning
information was destroyed. A \(D\)-dimensional state necessarily has
only \(D\) local degrees of freedom. The remaining logit-space
directions may be incompatible across examples, not representable by the
shared model, useful through other parameters, or unimportant for
reducing future loss.

This leaves two separate experimental questions:

\begin{enumerate}
\def\labelenumi{\arabic{enumi}.}
\tightlist
\item
  If we keep the usual logits and LM-head update, but remove some
  directions only from the gradient sent into the Transformer, how much
  does training degrade?
\item
  If we keep standard backpropagation intact, can we compress the
  vocabulary-space residual, inject it into earlier Transformer layers
  and improve training?
\end{enumerate}

The first question gives a clean causal control for the original
paper\textquotesingle s low-rank-head experiment. We compare the
backward-only intervention with a low-rank forward head, separating harm
caused by restricting the hidden-state gradient from harm caused by
changing the decoder itself.

The second question tests the proposed remedy more directly. We tried
fixed, adaptive, residual, multi-depth and meta-learned routes for the
residual. None reliably improved on tuned standard backpropagation.
These experiments cover several plausible encodings, but not every
possible one: there is no unique way to map a high-dimensional
vocabulary residual into an earlier hidden state.

Our experiments do not establish that the natural projection through the
LM head is a harmful bottleneck. They show that deliberately removing
directions that already reach the hidden state is harmful. They also
show that reducing the rank of the forward decoder is more harmful
still. The part that does not reach the current hidden state is not
discarded by the whole model: it trains the LM head itself. Other
evidence used to support the bottleneck---especially a repeated-token
synthetic task---also has simpler explanations.

\Needspace{8\baselineskip}

\section{Main Findings}\label{main-findings}

The experiments produced six main findings:

\begin{enumerate}
\def\labelenumi{\arabic{enumi}.}
\tightlist
\item
  \textbf{The severe projection is real.} When the vocabulary is much
  larger than the hidden state, most of the logit-gradient norm can lie
  outside the LM head\textquotesingle s current feedback directions.
\item
  \textbf{The part outside those directions is not lost to the whole
  model.} It updates the LM-head matrix and changes the directions
  available on later training steps. Removing this update is very
  harmful.
\item
  \textbf{Deleting directions that already reach the hidden state
  hurts.} This result appears in both a byte-level language model and a
  larger subword model. It does not show that the unavailable
  vocabulary-space directions would help if added to the model.
\item
  \textbf{Changing the forward decoder hurts much more than changing
  only the backward path.} A low-rank forward-head experiment therefore
  does not isolate a backward bottleneck.
\item
  \textbf{SpamLang confounds token count with independent data
  coverage.} The task samples one symbol, repeats it across the whole
  sequence, and trains the model to predict that same symbol. A 64-token
  sequence therefore supplies 64 loss terms, but only one independently
  sampled symbol sequence. When we held independent sequences per symbol
  fixed, increasing repetition did not change learning.
\item
  \textbf{The geometric measurements do not reliably predict learning
  progress.} They describe the projection, but we did not find strong
  evidence that they identify which models will learn faster.
\end{enumerate}

Several attempts to add a separate feedback route also failed to beat
ordinary, well-tuned backpropagation. Together, these results leave the
proposed causal problem unproven.

\Needspace{8\baselineskip}

\section{What ``Lost'' Means Here}\label{what-lost-means-here}

Let \(P_W\) be the orthogonal projector onto the column space of \(W\).
We can decompose the logit gradient as:

\[
g_z = P_Wg_z + (I-P_W)g_z
= g_{\parallel} + g_{\perp}.
\]

The orthogonal component \(g_{\perp}\) is invisible to the current
hidden state:

\[
W^T g_{\perp} = 0.
\]

But it is not invisible to the LM-head parameters. Their gradient is:

\[
\frac{\partial L}{\partial W} = g_z h^T.
\]

The component \(g_{\perp}\) contributes:

\[
g_{\perp}h^T.
\]

This term is generally nonzero. It changes \(W\), which changes both
future predictions and the feedback subspace available to later
examples.

The component is therefore invisible to the current hidden state, but
not discarded by the training process. A local projection is not the
same as information being permanently destroyed.

\Needspace{8\baselineskip}

\section{A Clean Test of Backward
Rank}\label{a-clean-test-of-backward-rank}

The original paper lowers the rank of the LM head and observes worse
training.

Lowering the rank of the forward head changes several things at once:

\begin{itemize}
\tightlist
\item
  which token-score patterns the model can produce;
\item
  the number and arrangement of trainable parameters;
\item
  the decoder\textquotesingle s conditioning; and
\item
  the rank of the signal sent backward.
\end{itemize}

The experiment cannot tell which change caused the loss.

We built a control that changes only the last item. The model computes
the ordinary full-rank logits. The LM head also receives its ordinary
parameter update. Only the signal returned to the hidden state is
replaced by a lower-rank version.

We used the best rank-\(r\) approximation of the current head and
matched the root-mean-square magnitude of the resulting hidden gradient
to the exact one. This prevents a change in gradient scale from being
mistaken for a rank effect. Each arm received its own learning-rate
search on seed 1 before confirmation. The final estimates use five
paired seeds. Seeds 4 and 5 were registered before their results were
inspected and reused every frozen hyperparameter.

We compared three cases:

\Needspace{8\baselineskip}\begingroup\small\sffamily\setlength{\tabcolsep}{6pt}\renewcommand{\arraystretch}{1}

{\def\LTcaptype{none} 
\begin{longtable}[]{@{}
  >{\raggedright\arraybackslash}p{(\linewidth - 4\tabcolsep) * \real{0.3333}}
  >{\raggedright\arraybackslash}p{(\linewidth - 4\tabcolsep) * \real{0.3333}}
  >{\raggedright\arraybackslash}p{(\linewidth - 4\tabcolsep) * \real{0.3333}}@{}}
\toprule\noalign{}\rowcolor{MathdownEditor}
\begin{minipage}[b]{\linewidth}\raggedright
\rule[-0.6em]{0pt}{2.8em}\bfseries\color{MathdownMuted}FORWARD
COMPUTATION
\end{minipage} & \begin{minipage}[b]{\linewidth}\raggedright
\bfseries\color{MathdownMuted}BACKWARD PATH
\end{minipage} & \begin{minipage}[b]{\linewidth}\raggedright
\bfseries\color{MathdownMuted}WHAT IT TESTS
\end{minipage} \\
\midrule\noalign{}
\endhead
\bottomrule\noalign{}
\endlastfoot
\rowcolor{MathdownEditor!24}\rule[-0.6em]{0pt}{2.8em}ordinary full-rank
head & exact gradient & normal training \\
\rowcolor{MathdownEditor!24}\rule[-0.6em]{0pt}{2.8em}ordinary full-rank
head & lower-rank gradient & backward rank only \\
\rowcolor{MathdownEditor!24}\rule[-0.6em]{0pt}{2.8em}lower-rank head &
its exact gradient & forward and backward changes together \\
\end{longtable}
}

\endgroup\par\noindent

The primary outcome is validation cross-entropy; lower is better. Tables
report the mean \(\pm\) standard error over five runs. For the main
causal differences, we also report two-sided 95\% Student-\(t\)
confidence intervals across paired seeds and Cohen\textquotesingle s
\(d_z\), the mean paired difference divided by its sample standard
deviation. Five seeds still provide limited precision; the intervals
should be read as uncertainty estimates, not asymptotic guarantees.
Because seed 1 was used for learning-rate selection, we also checked
every comparison using only seeds 2--5.

\Needspace{8\baselineskip}

\subsection{Relation to the Original
Experiment}\label{relation-to-the-original-experiment}

The original paper\textquotesingle s main pretraining comparison uses a
shared six-layer, width-4096 Transformer with about two billion
parameters and 11 billion training tokens. It varies a factorized
forward head from rank 32 to rank 4096 with a 49,152-token vocabulary.
The paper denotes this head rank by \(D\), while the Transformer width
remains fixed at 4096.

Our experiment is not a scale reproduction of that training run. Our
larger model has six layers, width 96, 2.26 million parameters, an
8,192-token vocabulary and 1.23 million supervised training tokens. Its
purpose is causal identification: for each retained rank, we compare the
original factorized forward intervention with a backward-only
intervention that leaves the logits and LM-head update unchanged.

\Needspace{8\baselineskip}\begingroup\small\sffamily\setlength{\tabcolsep}{6pt}\renewcommand{\arraystretch}{1}

{\def\LTcaptype{none} 
\begin{longtable}[]{@{}
  >{\raggedright\arraybackslash}p{(\linewidth - 4\tabcolsep) * \real{0.2800}}
  >{\raggedleft\arraybackslash}p{(\linewidth - 4\tabcolsep) * \real{0.3600}}
  >{\raggedleft\arraybackslash}p{(\linewidth - 4\tabcolsep) * \real{0.3600}}@{}}
\toprule\noalign{}\rowcolor{MathdownEditor}
\begin{minipage}[b]{\linewidth}\raggedright
\rule[-0.6em]{0pt}{2.8em}\bfseries\color{MathdownMuted}PROPERTY
\end{minipage} & \begin{minipage}[b]{\linewidth}\raggedleft
\bfseries\color{MathdownMuted}ORIGINAL PRETRAINING COMPARISON
\end{minipage} & \begin{minipage}[b]{\linewidth}\raggedleft
\bfseries\color{MathdownMuted}OUR LARGER CAUSAL CONTROL
\end{minipage} \\
\midrule\noalign{}
\endhead
\bottomrule\noalign{}
\endlastfoot
\rowcolor{MathdownEditor!24}\rule[-0.6em]{0pt}{2.8em}Transformer layers
& 6 & 6 \\
\rowcolor{MathdownEditor!24}\rule[-0.6em]{0pt}{2.8em}Transformer width &
4,096 & 96 \\
\rowcolor{MathdownEditor!24}\rule[-0.6em]{0pt}{2.8em}Parameters & about
2B & 2.26M \\
\rowcolor{MathdownEditor!24}\rule[-0.6em]{0pt}{2.8em}Vocabulary & 49,152
& 8,192 \\
\rowcolor{MathdownEditor!24}\rule[-0.6em]{0pt}{2.8em}Training tokens &
about 11B & 1.23M per run \\
\rowcolor{MathdownEditor!24}\rule[-0.6em]{0pt}{2.8em}Intervention &
factorized forward head & factorized forward and backward-only \\
\rowcolor{MathdownEditor!24}\rule[-0.6em]{0pt}{2.8em}Paired seeds
reported here & --- & 5 \\
\end{longtable}
}

\endgroup\par\noindent

The original experiment establishes that lower-rank forward heads train
worse at large scale. It does not isolate whether that loss comes from
the backward channel, the decoder\textquotesingle s forward
parameterization, or both. Our experiment isolates that distinction at
small scale; it does not establish that the same effect sizes hold for
billion-parameter models.

\Needspace{8\baselineskip}

\section{Compact Byte-Level Language
Model}\label{compact-byte-level-language-model}

The first experiment uses a byte-level WikiText-2 model with four
Transformer layers, hidden width 32 and 600 training steps.

\Needspace{8\baselineskip}\begingroup\small\sffamily\setlength{\tabcolsep}{6pt}\renewcommand{\arraystretch}{1}

{\def\LTcaptype{none} 
\begin{longtable}[]{@{}
  >{\raggedright\arraybackslash}p{(\linewidth - 6\tabcolsep) * \real{0.3700}}
  >{\raggedleft\arraybackslash}p{(\linewidth - 6\tabcolsep) * \real{0.1700}}
  >{\raggedleft\arraybackslash}p{(\linewidth - 6\tabcolsep) * \real{0.2200}}
  >{\raggedleft\arraybackslash}p{(\linewidth - 6\tabcolsep) * \real{0.2400}}@{}}
\toprule\noalign{}\rowcolor{MathdownEditor}
\begin{minipage}[b]{\linewidth}\raggedright
\rule[-0.6em]{0pt}{2.8em}\bfseries\color{MathdownMuted}FORWARD
COMPUTATION
\end{minipage} & \begin{minipage}[b]{\linewidth}\raggedleft
\bfseries\color{MathdownMuted}BACKWARD RANK
\end{minipage} & \begin{minipage}[b]{\linewidth}\raggedleft
\bfseries\color{MathdownMuted}VALIDATION LOSS
\end{minipage} & \begin{minipage}[b]{\linewidth}\raggedleft
\bfseries\color{MathdownMuted}INCREASE OVER ORDINARY TRAINING
\end{minipage} \\
\midrule\noalign{}
\endhead
\bottomrule\noalign{}
\endlastfoot
\rowcolor{MathdownEditor!24}\rule[-0.6em]{0pt}{2.8em}ordinary head & 32
& \(2.2530 \pm 0.0192\) & --- \\
\rowcolor{MathdownEditor!24}\rule[-0.6em]{0pt}{2.8em}ordinary head & 16
& \(2.3164 \pm 0.0158\) & \(+0.0634\) \\
\rowcolor{MathdownEditor!24}\rule[-0.6em]{0pt}{2.8em}ordinary head & 8 &
\(2.3333 \pm 0.0123\) & \(+0.0804\) \\
\rowcolor{MathdownEditor!24}\rule[-0.6em]{0pt}{2.8em}ordinary head & 4 &
\(2.4545 \pm 0.0052\) & \(+0.2016\) \\
\rowcolor{MathdownEditor!24}\rule[-0.6em]{0pt}{2.8em}rank-16 forward
head & 16 & \(2.3972 \pm 0.0125\) & \(+0.1443\) \\
\rowcolor{MathdownEditor!24}\rule[-0.6em]{0pt}{2.8em}rank-8 forward head
& 8 & \(2.4363 \pm 0.0096\) & \(+0.1834\) \\
\rowcolor{MathdownEditor!24}\rule[-0.6em]{0pt}{2.8em}rank-4 forward head
& 4 & \(2.5458 \pm 0.0120\) & \(+0.2928\) \\
\end{longtable}
}

\endgroup\par\noindent

Removing directions from the backward path made training worse. The
effect grew as more directions were removed.

But changing the forward head was consistently worse than changing only
the backward path. The direct paired difference between the forward and
backward-only interventions was \(+0.0808\) at rank 16, \(+0.1030\) at
rank 8 and \(+0.0913\) at rank 4. All 15 paired comparisons had the
predicted sign, and the 95\% intervals for the three differences
excluded zero.

\Needspace{8\baselineskip}\begingroup\small\sffamily\setlength{\tabcolsep}{6pt}\renewcommand{\arraystretch}{1}

{\def\LTcaptype{none} 
\begin{longtable}[]{@{}
  >{\raggedleft\arraybackslash}p{(\linewidth - 8\tabcolsep) * \real{0.0800}}
  >{\raggedleft\arraybackslash}p{(\linewidth - 8\tabcolsep) * \real{0.3400}}
  >{\raggedleft\arraybackslash}p{(\linewidth - 8\tabcolsep) * \real{0.0800}}
  >{\raggedleft\arraybackslash}p{(\linewidth - 8\tabcolsep) * \real{0.4200}}
  >{\raggedleft\arraybackslash}p{(\linewidth - 8\tabcolsep) * \real{0.0800}}@{}}
\toprule\noalign{}\rowcolor{MathdownEditor}
\begin{minipage}[b]{\linewidth}\raggedleft
\rule[-0.6em]{0pt}{2.8em}\bfseries\color{MathdownMuted}RANK
\end{minipage} & \begin{minipage}[b]{\linewidth}\raggedleft
\bfseries\color{MathdownMuted}BACKWARD-ONLY MINUS EXACT, 95\% CI
\end{minipage} & \begin{minipage}[b]{\linewidth}\raggedleft
\bfseries\color{MathdownMuted}\(d_z\)
\end{minipage} & \begin{minipage}[b]{\linewidth}\raggedleft
\bfseries\color{MathdownMuted}FORWARD MINUS BACKWARD-ONLY, 95\% CI
\end{minipage} & \begin{minipage}[b]{\linewidth}\raggedleft
\bfseries\color{MathdownMuted}\(d_z\)
\end{minipage} \\
\midrule\noalign{}
\endhead
\bottomrule\noalign{}
\endlastfoot
\rowcolor{MathdownEditor!24}\rule[-0.6em]{0pt}{2.8em}16 &
\(+0.0634\ [0.0413,0.0856]\) & \(3.55\) & \(+0.0808\ [0.0303,0.1313]\) &
\(1.99\) \\
\rowcolor{MathdownEditor!24}\rule[-0.6em]{0pt}{2.8em}8 &
\(+0.0804\ [0.0503,0.1104]\) & \(3.32\) & \(+0.1030\ [0.0538,0.1522]\) &
\(2.60\) \\
\rowcolor{MathdownEditor!24}\rule[-0.6em]{0pt}{2.8em}4 &
\(+0.2016\ [0.1469,0.2562]\) & \(4.58\) & \(+0.0913\ [0.0569,0.1256]\) &
\(3.30\) \\
\end{longtable}
}

\endgroup\par\noindent

The rank-32 backward control reproduced ordinary backpropagation
exactly. This checked that the custom backward operation did not change
the forward pass or the LM head\textquotesingle s own update.

\begin{center}
\fcolorbox{MathdownLine}{white}{\includegraphics[width=0.965\linewidth]{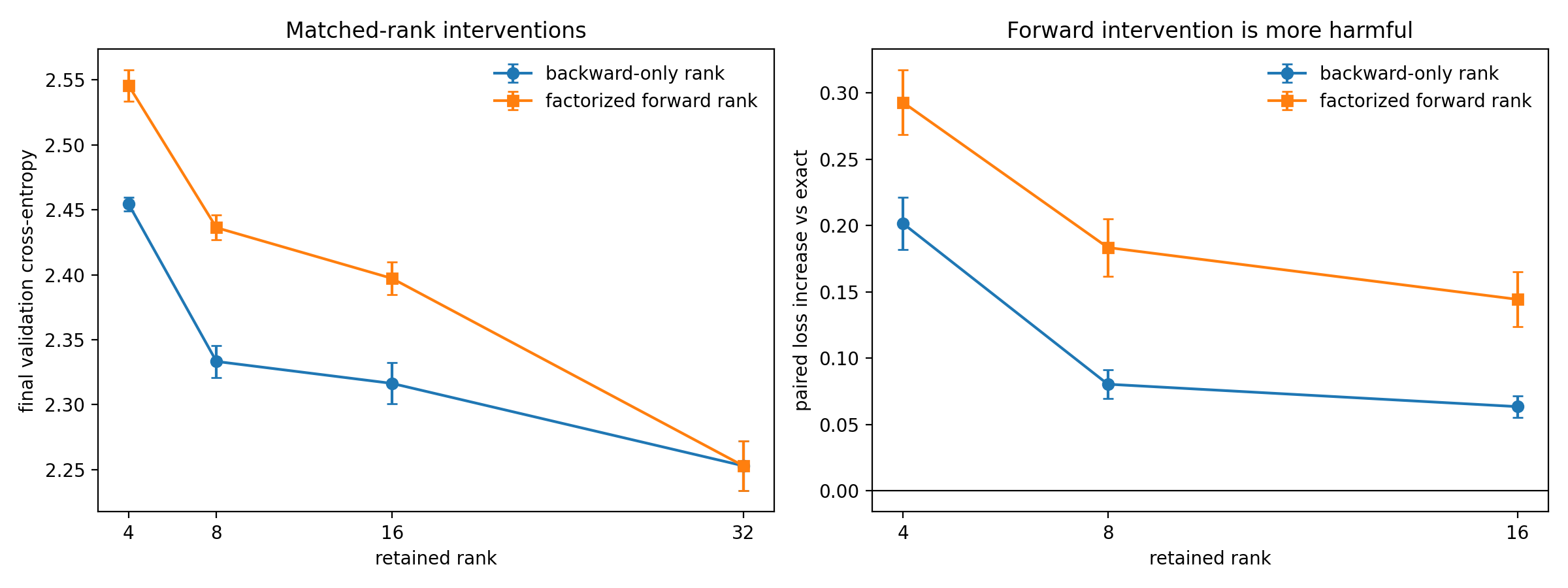}}
\end{center}

This experiment shows that directions already present in the hidden
gradient matter. It does not show that directions outside this
\(32\)-dimensional space would help the model. Those are different
claims.

\Needspace{8\baselineskip}

\section{Larger Subword Language
Model}\label{larger-subword-language-model}

We repeated the main comparison with a BPE-8192 model with six
Transformer layers, hidden width 96, about 2.26 million parameters and
1,200 training steps.

\Needspace{8\baselineskip}\begingroup\small\sffamily\setlength{\tabcolsep}{6pt}\renewcommand{\arraystretch}{1}

{\def\LTcaptype{none} 
\begin{longtable}[]{@{}
  >{\raggedright\arraybackslash}p{(\linewidth - 6\tabcolsep) * \real{0.3700}}
  >{\raggedleft\arraybackslash}p{(\linewidth - 6\tabcolsep) * \real{0.1700}}
  >{\raggedleft\arraybackslash}p{(\linewidth - 6\tabcolsep) * \real{0.2200}}
  >{\raggedleft\arraybackslash}p{(\linewidth - 6\tabcolsep) * \real{0.2400}}@{}}
\toprule\noalign{}\rowcolor{MathdownEditor}
\begin{minipage}[b]{\linewidth}\raggedright
\rule[-0.6em]{0pt}{2.8em}\bfseries\color{MathdownMuted}FORWARD
COMPUTATION
\end{minipage} & \begin{minipage}[b]{\linewidth}\raggedleft
\bfseries\color{MathdownMuted}RANK
\end{minipage} & \begin{minipage}[b]{\linewidth}\raggedleft
\bfseries\color{MathdownMuted}VALIDATION LOSS
\end{minipage} & \begin{minipage}[b]{\linewidth}\raggedleft
\bfseries\color{MathdownMuted}INCREASE OVER ORDINARY TRAINING
\end{minipage} \\
\midrule\noalign{}
\endhead
\bottomrule\noalign{}
\endlastfoot
\rowcolor{MathdownEditor!24}\rule[-0.6em]{0pt}{2.8em}ordinary head and
exact gradient & 96 & \(5.9692 \pm 0.0230\) & --- \\
\rowcolor{MathdownEditor!24}\rule[-0.6em]{0pt}{2.8em}ordinary head and
lower-rank gradient & 48 & \(6.0278 \pm 0.0201\) & \(+0.0586\) \\
\rowcolor{MathdownEditor!24}\rule[-0.6em]{0pt}{2.8em}rank-48 forward
head and exact gradient & 48 & \(6.1487 \pm 0.0190\) & \(+0.1795\) \\
\end{longtable}
}

\endgroup\par\noindent

The result followed the same pattern. Removing half the directions from
the backward path caused a modest loss increase: \(+0.0586\), with 95\%
CI \([+0.0167,+0.1005]\) and paired \(d_z=1.74\). Reducing the forward
rank caused an increase of \(+0.1795\) (\([+0.1547,+0.2042]\),
\(d_z=9.00\)). The forward head was worse than the backward-only
intervention by \(+0.1209\) (\([+0.0848,+0.1569]\), \(d_z=4.16\)). Every
paired difference had the predicted sign across all five seeds.

The four-seed sensitivity analysis that excludes the learning-rate
screening seed preserves the sign and ordering of every comparison. Its
95\% interval for forward minus backward-only remains above zero:
\([+0.0744,+0.1756]\). The smaller backward-only penalty has a wider
interval of \([-0.0046,+0.1108]\). We therefore regard the
forward-versus-backward distinction as the stronger result and the
standalone half-rank backward penalty as less precisely estimated in
this model.

For practical reasons, the lower-rank backward matrix was recalculated
every 20 training steps in this experiment rather than every step. It
was computed from a \(96 \times 96\) matrix, not by decomposing the full
vocabulary matrix.

\begin{center}
\fcolorbox{MathdownLine}{white}{\includegraphics[width=0.965\linewidth]{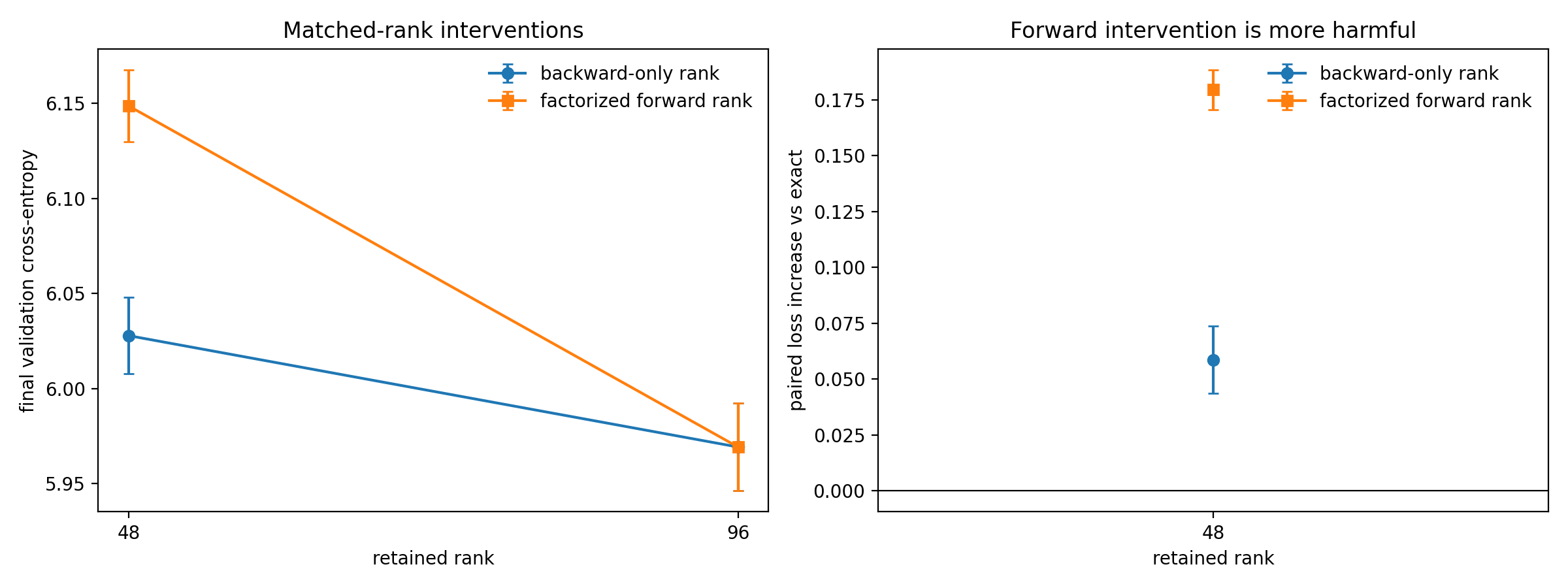}}
\end{center}

The larger experiment supports a narrow conclusion: deleting an existing
hidden-gradient direction can be harmful. It still does not show that
the much larger logit-space residual contains an improvement that the
shared model could use.

\Needspace{8\baselineskip}

\section{Is the Remaining Logit Error Already
Used?}\label{is-the-remaining-logit-error-already-used}

We next tested the part of the logit error that does not reach the
current hidden state.

Ordinary training uses this part to update the LM head. We removed it
only from the head update while keeping the exact hidden-state gradient.
We again matched the update size and searched the learning rate
separately. This secondary comparison retains the original three paired
seeds.

\Needspace{8\baselineskip}\begingroup\small\sffamily\setlength{\tabcolsep}{6pt}\renewcommand{\arraystretch}{1}

{\def\LTcaptype{none} 
\begin{longtable}[]{@{}
  >{\raggedright\arraybackslash}p{(\linewidth - 4\tabcolsep) * \real{0.3333}}
  >{\raggedleft\arraybackslash}p{(\linewidth - 4\tabcolsep) * \real{0.3333}}
  >{\raggedleft\arraybackslash}p{(\linewidth - 4\tabcolsep) * \real{0.3333}}@{}}
\toprule\noalign{}\rowcolor{MathdownEditor}
\begin{minipage}[b]{\linewidth}\raggedright
\rule[-0.6em]{0pt}{2.8em}\bfseries\color{MathdownMuted}HEAD UPDATE
\end{minipage} & \begin{minipage}[b]{\linewidth}\raggedleft
\bfseries\color{MathdownMuted}VALIDATION LOSS
\end{minipage} & \begin{minipage}[b]{\linewidth}\raggedleft
\bfseries\color{MathdownMuted}INCREASE OVER ORDINARY TRAINING
\end{minipage} \\
\midrule\noalign{}
\endhead
\bottomrule\noalign{}
\endlastfoot
\rowcolor{MathdownEditor!24}\rule[-0.6em]{0pt}{2.8em}complete logit
gradient & \(2.2621 \pm 0.0321\) & --- \\
\rowcolor{MathdownEditor!24}\rule[-0.6em]{0pt}{2.8em}projected logit
gradient only & \(2.8873 \pm 0.0536\) & \(+0.6252\) \\
\end{longtable}
}

\endgroup\par\noindent

Removing the orthogonal component from the head update was the most
damaging intervention in the compact study.

This result does not show that the component should also be sent into
the Transformer. It shows something simpler: ordinary training already
uses it. The component changes the token vectors in the head and rotates
the feedback directions available to later examples. Describing its full
length as destroyed by the model is therefore inaccurate.

\Needspace{8\baselineskip}\section{SpamLang: Repeated Tokens Are\protect\linebreak Not Independent Mapping Examples}\label{spamlang-repeated-tokens-are-not-independent-mapping-examples}

The original paper also studies a synthetic task called SpamLang. To
construct one training sequence, it:

\begin{enumerate}
\def\labelenumi{\arabic{enumi}.}
\tightlist
\item
  samples one vocabulary symbol \(x\); and
\item
  repeats \(x\) at every position in the sequence.
\end{enumerate}

The language model must predict the next token, which is also \(x\). For
sequence length \(L\), one example is therefore:

\[
\text{input}=(x,x,\ldots,x), \qquad
\text{target}=(x,x,\ldots,x).
\]

The model is therefore learning a token-identity relation over the
vocabulary. A larger vocabulary creates more token identities to learn.

The paper counts every position as another occurrence of the symbol.
That count is correct as a count of supervised tokens, but not as a
count of independently sampled symbol sequences. All positions in one
sequence repeat the same identity relation \(x \mapsto x\).

For example, a length-64 sequence supplies 64 cross-entropy terms for
the same token, but it does not expose the model to 64 independently
sampled symbols.

For our controlled exposure test, we replaced the identity target with a
fixed shuffled mapping \(x \mapsto \pi(x)\). This removes the shortcut
of copying the input token while preserving SpamLang\textquotesingle s
sampling and repetition pattern. In the embedding-plus-linear model used
for this test, the 64 repeated terms produce copies of the same gradient
contribution, apart from loss scaling.

First, we fixed the number of independently drawn examples at four per
symbol:

\Needspace{8\baselineskip}\begingroup\small\sffamily\setlength{\tabcolsep}{6pt}\renewcommand{\arraystretch}{1}

{\def\LTcaptype{none} 
\begin{longtable}[]{@{}
  >{\raggedleft\arraybackslash}p{(\linewidth - 6\tabcolsep) * \real{0.1800}}
  >{\raggedleft\arraybackslash}p{(\linewidth - 6\tabcolsep) * \real{0.3000}}
  >{\raggedleft\arraybackslash}p{(\linewidth - 6\tabcolsep) * \real{0.2800}}
  >{\raggedleft\arraybackslash}p{(\linewidth - 6\tabcolsep) * \real{0.2400}}@{}}
\toprule\noalign{}\rowcolor{MathdownEditor}
\begin{minipage}[b]{\linewidth}\raggedleft
\rule[-0.6em]{0pt}{2.8em}\bfseries\color{MathdownMuted}SEQUENCE LENGTH
\end{minipage} & \begin{minipage}[b]{\linewidth}\raggedleft
\bfseries\color{MathdownMuted}INDEPENDENT DRAWS PER SYMBOL
\end{minipage} & \begin{minipage}[b]{\linewidth}\raggedleft
\bfseries\color{MathdownMuted}COUNTED TOKEN OCCURRENCES
\end{minipage} & \begin{minipage}[b]{\linewidth}\raggedleft
\bfseries\color{MathdownMuted}VALIDATION LOSS
\end{minipage} \\
\midrule\noalign{}
\endhead
\bottomrule\noalign{}
\endlastfoot
\rowcolor{MathdownEditor!24}\rule[-0.6em]{0pt}{2.8em}1 & 4 & 4 &
\(4.4455 \pm 0.0294\) \\
\rowcolor{MathdownEditor!24}\rule[-0.6em]{0pt}{2.8em}64 & 4 & 256 &
\(4.4455 \pm 0.0294\) \\
\end{longtable}
}

\endgroup\par\noindent

The two five-run results were the same. Within each matching seed, their
training paths agreed to numerical precision. Repetition increased the
number of supervised positions from 4 to 256 per symbol without adding
another independently sampled mapping example.

We then fixed the total number of supervised token positions at 131,072:

\Needspace{8\baselineskip}\begingroup\small\sffamily\setlength{\tabcolsep}{6pt}\renewcommand{\arraystretch}{1}

{\def\LTcaptype{none} 
\begin{longtable}[]{@{}
  >{\raggedleft\arraybackslash}p{(\linewidth - 8\tabcolsep) * \real{0.1600}}
  >{\raggedleft\arraybackslash}p{(\linewidth - 8\tabcolsep) * \real{0.2500}}
  >{\raggedleft\arraybackslash}p{(\linewidth - 8\tabcolsep) * \real{0.2300}}
  >{\raggedleft\arraybackslash}p{(\linewidth - 8\tabcolsep) * \real{0.2000}}
  >{\raggedleft\arraybackslash}p{(\linewidth - 8\tabcolsep) * \real{0.1600}}@{}}
\toprule\noalign{}\rowcolor{MathdownEditor}
\begin{minipage}[b]{\linewidth}\raggedleft
\rule[-0.6em]{0pt}{2.8em}\bfseries\color{MathdownMuted}SEQUENCE LENGTH
\end{minipage} & \begin{minipage}[b]{\linewidth}\raggedleft
\bfseries\color{MathdownMuted}INDEPENDENT DRAWS PER SYMBOL
\end{minipage} & \begin{minipage}[b]{\linewidth}\raggedleft
\bfseries\color{MathdownMuted}COUNTED TOKEN OCCURRENCES
\end{minipage} & \begin{minipage}[b]{\linewidth}\raggedleft
\bfseries\color{MathdownMuted}VALIDATION LOSS
\end{minipage} & \begin{minipage}[b]{\linewidth}\raggedleft
\bfseries\color{MathdownMuted}ACCURACY
\end{minipage} \\
\midrule\noalign{}
\endhead
\bottomrule\noalign{}
\endlastfoot
\rowcolor{MathdownEditor!24}\rule[-0.6em]{0pt}{2.8em}1 & 128 & 128 &
\(0.0296 \pm 0.0032\) & \(100.0\%\) \\
\rowcolor{MathdownEditor!24}\rule[-0.6em]{0pt}{2.8em}64 & 2 & 128 &
\(6.1229 \pm 0.0359\) & \(26.6\%\) \\
\end{longtable}
}

\endgroup\par\noindent

Both conditions reported the same 128 supervised positions per symbol.
With sequence length 1, those positions came from 128 independently
sampled sequences. With sequence length 64, they came from only two
sequences. The first condition solved the task; the second did not.

At the original paper\textquotesingle s largest setting, 41 million
token positions, sequence length 64 and vocabulary size 131,072 provide
only about 4.9 independently sampled sequences per symbol:

\[
\frac{41{,}000{,}000}{64 \times 131{,}072} \approx 4.9.
\]

The reported figure of roughly 300 occurrences per symbol counts the
repeated positions. It does not mean that the model saw roughly 300
independent examples of each mapping.

The large-vocabulary setting therefore changes more than the proposed
gradient geometry: it leaves very few independently sampled sequences
for each symbol. This is a direct alternative explanation for the
failure.

Repeated positions can still affect a contextual
Transformer\textquotesingle s activations and optimization, so they are
not always computationally equivalent to one position. The narrower
point is statistical: they do not provide independent coverage of the
symbol-to-symbol mapping. SpamLang must control this coverage before its
failure can be attributed to the LM-head gradient path.

\Needspace{8\baselineskip}

\section{Increasing the Output Vocabulary Without Changing the
Text}\label{increasing-the-output-vocabulary-without-changing-the-text}

Changing vocabulary size usually changes tokenization, token frequency,
sequence length and the number of examples seen for each token. These
changes make it difficult to isolate the effect of \(V/D\).

We used a simpler control. The input remained byte-level text with the
same 256 possible input bytes. We added output classes that were never
correct targets. This increased the size of the softmax while leaving
the text, targets, input embeddings, Transformer, batch order and number
of training tokens unchanged.

\Needspace{8\baselineskip}\begingroup\small\sffamily\setlength{\tabcolsep}{6pt}\renewcommand{\arraystretch}{1}

{\def\LTcaptype{none} 
\begin{longtable}[]{@{}
  >{\raggedleft\arraybackslash}p{(\linewidth - 4\tabcolsep) * \real{0.3333}}
  >{\raggedleft\arraybackslash}p{(\linewidth - 4\tabcolsep) * \real{0.3333}}
  >{\raggedleft\arraybackslash}p{(\linewidth - 4\tabcolsep) * \real{0.3333}}@{}}
\toprule\noalign{}\rowcolor{MathdownEditor}
\begin{minipage}[b]{\linewidth}\raggedleft
\rule[-0.6em]{0pt}{2.8em}\bfseries\color{MathdownMuted}OUTPUT CLASSES
\end{minipage} & \begin{minipage}[b]{\linewidth}\raggedleft
\bfseries\color{MathdownMuted}HIDDEN WIDTH
\end{minipage} & \begin{minipage}[b]{\linewidth}\raggedleft
\bfseries\color{MathdownMuted}VALIDATION LOSS
\end{minipage} \\
\midrule\noalign{}
\endhead
\bottomrule\noalign{}
\endlastfoot
\rowcolor{MathdownEditor!24}\rule[-0.6em]{0pt}{2.8em}256 & 32 &
\(2.2591 \pm 0.0296\) \\
\rowcolor{MathdownEditor!24}\rule[-0.6em]{0pt}{2.8em}1,024 & 32 &
\(2.2532 \pm 0.0161\) \\
\rowcolor{MathdownEditor!24}\rule[-0.6em]{0pt}{2.8em}4,096 & 32 &
\(2.2642 \pm 0.0110\) \\
\end{longtable}
}

\endgroup\par\noindent

Increasing \(V/D\) sixteenfold did not make the tuned model worse. A
second test with hidden width 64 gave the same result for one seed: loss
was 2.2336 with 256 output classes and 2.2322 with 4,096 output classes.

Never-correct output classes are not the same as a natural larger
tokenizer. This experiment does not show that vocabulary size never
matters. It shows that the number of output choices, and the competition
among those choices, did not create the predicted optimization problem
in this controlled setting.

\Needspace{8\baselineskip}

\section{Do Projection Measurements Predict
Learning?}\label{do-projection-measurements-predict-learning}

A useful bottleneck measurement should do more than change when \(V/D\)
changes. It should help predict which training runs will improve.

After the five-seed extension, we collected measurements from 894 usable
saved training points across 37 experimental conditions. These included:

\begin{itemize}
\tightlist
\item
  how much logit-gradient norm survives the projection;
\item
  the angle between the original and projected signals;
\item
  how well the target token and strongest competing token are preserved;
\item
  the number and strength of the logit gradient\textquotesingle s main
  directions; and
\item
  the condition number of the LM head, which measures numerical
  imbalance across directions.
\end{itemize}

We first predicted learning from ordinary information such as current
loss, training progress, learning rate, model width, dataset and
experimental method. We then added one projection measurement and tested
on an experimental condition left out during fitting.

Adding retained gradient length reduced the prediction error for final
loss by 8.2\%, but its 95\% uncertainty interval ranged from a 23.0\%
improvement to a 3.4\% worsening. For the next interval of learning, the
estimated improvement was 3.1\%, with an interval from 11.7\% better to
5.0\% worse. The other measurements were similarly uncertain or
unhelpful.

These results do not prove that the measurements contain no useful
information. They show that, in our runs, none reliably predicted
short-term learning once basic experimental differences were taken into
account.

\Needspace{8\baselineskip}

\section{Secondary Evidence: Tested Auxiliary Feedback
Routes}\label{secondary-evidence-tested-auxiliary-feedback-routes}

These experiments are secondary to the backward-only causal test. We
also tested the direct engineering idea suggested by the bottleneck
claim: keep or replace the ordinary LM-head gradient with a separately
designed return path.

The tested methods included:

\begin{itemize}
\tightlist
\item
  a fixed random return matrix;
\item
  a matrix that adapted to the largest recent logit-gradient directions;
\item
  an extra route carrying the part not returned by the ordinary head;
\item
  several routes entering different Transformer depths;
\item
  extra prediction losses at intermediate layers; and
\item
  a one-step meta-learned route optimized on a fresh batch.
\end{itemize}

Some methods learned successfully. Some helped when the ordinary
model\textquotesingle s learning rate was poorly chosen. None gave a
reliable improvement over ordinary backpropagation after learning rates
were tuned and results were repeated. The meta-learned route produced a
small one-step gain, but a 10\% increase in the ordinary SGD learning
rate produced a larger gain. It therefore did not isolate a useful new
source of credit.

The adaptive method is especially instructive. It tracked directions
with the largest recent squared gradient. Those are the directions that
best reconstruct the logit gradient under Euclidean norm. But high
variance in logit space does not imply a useful descent direction after
the signal is mapped back to shared model parameters. Tracking the
dominant modes therefore made mathematical sense without producing
better optimization.

This is a negative result for the methods we tested, not for every
possible feedback design. A different architecture or learning rule
could still help.

\Needspace{8\baselineskip}

\section{What the Experiments
Establish}\label{what-the-experiments-establish}

The evidence supports the following statements:

\begin{enumerate}
\def\labelenumi{\arabic{enumi}.}
\tightlist
\item
  logit gradients collected across a batch can span many more directions
  than the hidden width;
\item
  only a small part of their Euclidean norm may reach the current final
  hidden state;
\item
  deleting directions that already reach that state makes training
  worse;
\item
  restricting the forward decoder is more damaging than an equal
  restriction applied only during the backward pass;
\item
  the remaining logit error strongly trains the LM head itself; and
\item
  repeated positions are not a substitute for independent symbol
  examples.
\end{enumerate}

The experiments do not establish these stronger statements:

\begin{enumerate}
\def\labelenumi{\arabic{enumi}.}
\tightlist
\item
  the full logit-gradient remainder is a useful update for the
  Transformer;
\item
  the Transformer could follow that remainder while preserving useful
  sharing across different pieces of text;
\item
  the LM head causes the large-vocabulary synthetic failure;
\item
  retained gradient length measures retained learning value; or
\item
  a wider or separate feedback path would improve ordinary
  language-model training.
\end{enumerate}

The distinction is between geometry and causation. The projection
exists. Its harm has not been demonstrated.

\Needspace{8\baselineskip}

\section{Relation to Earlier Work}\label{relation-to-earlier-work}

Other research has shown that neural networks can learn with approximate
return signals.

\textbf{Feedback alignment} replaces the exact transpose of each forward
matrix with a fixed random matrix. \textbf{Direct feedback alignment}
sends output errors directly to earlier layers. These results show that
exact symmetry is not always required for learning. They do not show
that an approximate signal is better than the exact gradient when the
exact gradient is available.

\textbf{Synthetic gradients} use a learned model to predict a later
gradient. Their main purpose is to let different parts of a network
train without waiting for one another. \textbf{Deep supervision} gives
intermediate layers their own prediction losses. This changes the
model\textquotesingle s training objective as well as its backward path.

Online methods such as \textbf{Oja\textquotesingle s rule} and
\textbf{Frequent Directions} track the main directions in a stream of
data. They are natural tools for compressing recent logit gradients.
Their goal is accurate reconstruction, however, not necessarily the best
future parameter update.

Methods such as \textbf{PowerSGD} compress parameter gradients for
distributed communication. This happens after the model has produced an
update that can be expressed in its parameter space. It is different
from treating an unconstrained vocabulary-space residual as supervision
that the shared model should follow.

\Needspace{8\baselineskip}

\section{Limits of This Study}\label{limits-of-this-study}

The largest model in this study has about 2.26 million parameters, six
layers and hidden width 96. It trained for 1.23 million supervised
tokens in the causal experiment. The results do not establish what
happens in models with billions of parameters or during long pretraining
runs.

The larger backward-only experiment recalculated its lower-rank
approximation every 20 steps. Updating it every step could produce a
different result.

The added-output-class experiment cleanly changes output dimension, but
it is not a natural tokenizer comparison. Natural tokenizers also change
the data seen by the model.

The prediction study contains many related small-model runs. Saved
training points from the same run are not independent observations, and
the uncertainty intervals remain wide.

Most importantly, failing to find a better feedback method does not
prove that none exists. The bottleneck claim would gain stronger support
if a route for the unavailable logit error kept the
model\textquotesingle s predictions unchanged yet reliably beat exact
backpropagation across learning rates, seeds, tasks and model sizes. Its
gain should also grow in a predictable way with the severity of the
measured projection.

\Needspace{8\baselineskip}

\section{Reproducing the Results}\label{reproducing-the-results}

The
\href{https://github.com/kupilikula/lm-head-gradient-bottleneck}{public
GitHub repository} includes the code, frozen configurations, data
recipe, tests, figures and filtered per-run aggregate results used in
this article. Prepared corpora, checkpoints, step-by-step logs and
machine environment captures are omitted because they are large
generated artifacts. A separate source manifest records the origin and
checksum of every included result table.

The main files are:

\begin{itemize}
\tightlist
\item
  frozen experimental choices: \texttt{docs/PROTOCOL\_LOCK.md};
\item
  commands and the result evidence map:
  \texttt{docs/REPRODUCIBILITY.md};
\item
  causal experiment runner: \texttt{src/bottleneck/response\_pilot.py};
\item
  synthetic exposure runner:
  \texttt{src/bottleneck/response\_synthetic.py};
\item
  controlled output-size runner:
  \texttt{src/bottleneck/response\_vocab.py};
\item
  prediction analysis: \texttt{src/bottleneck/response\_predictive.py};
  and
\item
  dataset sources and checksums: \texttt{data/README.md}.
\end{itemize}

The release verifier checks result checksums and scans for accidental
local paths, credentials, corpora and checkpoints:

\begin{Shaded}
\begin{Highlighting}[]
\ExtensionTok{uv}\NormalTok{ run python scripts/verify\_release.py}
\end{Highlighting}
\end{Shaded}

The test suite checks that the forward computation stays unchanged in
the backward-only intervention, that full-rank custom feedback matches
ordinary backpropagation, and that data generation is repeatable.

\Needspace{8\baselineskip}

\section{Conclusion}\label{conclusion}

The LM head maps a small internal state to a much larger vocabulary. Its
return path can therefore carry only a small part of an arbitrary
vocabulary-space error. This is a real and sometimes severe projection.

But a severe projection is not enough to establish a harmful bottleneck.

The part that does not reach the current hidden state still trains the
LM head. Reducing only the backward rank hurts, but reducing the forward
decoder at the same rank hurts much more. The repeated-token task is
largely explained by how few independent symbols the model observes.
Increasing output dimension alone does not reproduce the claimed
failure. Measurements of retained gradient length do not reliably
predict learning progress, and the separate feedback routes we tested do
not improve tuned training.

The current evidence therefore supports a limited conclusion: the LM
head strongly compresses the immediate error returned to the final
hidden state, but it has not been shown to discard useful learning
information in a way that limits language-model optimization or
performance.

That leaves a clear standard for future work. A convincing bottleneck
result should hold the forward model fixed, provide a route for the
missing signal that produces an update expressible in the
model\textquotesingle s parameter space, improve actual training rather
than only a geometric score, and repeat across model sizes and tasks.

\Needspace{8\baselineskip}

\section{Acknowledgements}\label{acknowledgements}

This project was directed by Anand Murugan and developed in
collaboration with OpenAI Codex using GPT-5.6-Sol. Codex contributed to
experimental design, implementation, statistical analysis, literature
review, reproducibility work and manuscript development. Anand Murugan
made the research decisions, reviewed the evidence and takes
responsibility for the claims and conclusions.

\Needspace{8\baselineskip}

\section{References}\label{references}

\begin{itemize}
\tightlist
\item
  Godey, N. and Artzi, Y.
  \href{https://arxiv.org/abs/2603.10145}{\emph{Lost in Backpropagation:
  The LM Head is a Gradient Bottleneck}}, 2026.
\item
  Lillicrap, T. et al.
  \href{https://arxiv.org/abs/1411.0247}{\emph{Random Synaptic Feedback
  Weights Support Error Backpropagation for Deep Learning}}.
\item
  Nøkland, A. \href{https://arxiv.org/abs/1609.01596}{\emph{Direct
  Feedback Alignment Provides Learning in Deep Neural Networks}}.
\item
  Jaderberg, M. et al.
  \href{https://arxiv.org/abs/1608.05343}{\emph{Decoupled Neural
  Interfaces Using Synthetic Gradients}}.
\item
  Lee, C.-Y. et al.
  \href{https://arxiv.org/abs/1409.5185}{\emph{Deeply-Supervised Nets}}.
\item
  Oja, E.
  \href{https://pubmed.ncbi.nlm.nih.gov/7153672/}{\emph{Simplified
  Neuron Model as a Principal Component Analyzer}}.
\item
  Ghashami, M. et al.
  \href{https://epubs.siam.org/doi/10.1137/15M1009718}{\emph{Frequent
  Directions: Simple and Deterministic Matrix Sketching}}.
\item
  Vogels, T. et al.
  \href{https://arxiv.org/abs/1905.13727}{\emph{PowerSGD: Practical
  Low-Rank Gradient Compression for Distributed Optimization}}.
\end{itemize}

\end{document}